\documentclass[a4, 10 pt, conference]{ieeeconf}

\IEEEoverridecommandlockouts                            
\usepackage[utf8]{inputenc}
\usepackage[T1]{fontenc}
\usepackage{graphicx}
\usepackage{xcolor}
\usepackage{amsmath, amssymb, amsfonts}
\usepackage{mathtools}
\usepackage{siunitx}
\usepackage{cite}
\usepackage{tikz}
\usepackage{bm}
\usepackage{microtype}
\usepackage[font=small]{caption}
\usepackage{subcaption}
\usepackage{booktabs}
\usepackage{tikz-3dplot}
\usepackage[hidelinks]{hyperref}

\let\labelindent\relax
\usepackage{enumitem}

\newcommand\copyrighttext{%
  \footnotesize \textcopyright 2026 IEEE. Personal use of this material is permitted.  Permission from IEEE must be obtained for all other uses, in any current or future media, including reprinting/republishing this material for advertising or promotional purposes, creating new collective works, for resale or redistribution to servers or lists, or reuse of any copyrighted component of this work in other works.}

\newcommand\copyrightnotice{%
\begin{tikzpicture}[remember picture,overlay]
\node[anchor=south,yshift=10pt] at (current page.south) {\parbox{\textwidth}{\copyrighttext}};
\end{tikzpicture}%
}

\makeatletter
\def\@IEEEsectpunct{.\ \,}
\makeatother

\newcommand{\uprightsuperscript}[1]{^{\textnormal{#1}}}
\begingroup\lccode`~=`*\lowercase{\endgroup\let~}\uprightsuperscript
\AtBeginDocument{\mathcode`*="8000 }

\newcommand{\uprightsubscript}[1]{_{\textnormal{#1}}}
\begingroup\lccode`~=`?\lowercase{\endgroup\let~}\uprightsubscript
\AtBeginDocument{\mathcode`?="8000 }

\definecolor{maroon}{rgb}{0.6350, 0.0780, 0.1840}
\definecolor{blue}{rgb}{0, 0.447, 0.741}
\definecolor{orange}{rgb}{0.85, 0.325, 0.098}
\definecolor{green}{rgb}{0.466, 0.674,0.188}
\definecolor{yellow}{rgb}{0.929, 0.694, 0.125}
\definecolor{light_gray}{rgb}{0.8, 0.8, 0.8}
\definecolor{dark_gray}{rgb}{0.149, 0.149, 0.149}

\newif\ifwithauthors
\withauthorstrue
\title{\LARGE \bf
        Nonlinear Predictive Control of the Continuum and Hybrid Dynamics of a Suspended Deformable Cable for Aerial Pick and Place
}

\newcommand{\myauthors}{%
    \author{Antonio Rapuano$^1$, Yaolei Shen$^2$, Federico Califano$^2$, Chiara Gabellieri$^2$ and Antonio Franchi$^{1,2}$%
    \thanks{
        This work was partially supported by the Horizon Europe research project 101120732 (AUTOASSESS) and the MSCA-SE project 101182891 (NEUTRAWEED). Views and opinions expressed are those of the authors and do not necessarily reflect those of the European Union or Research Executive Agency. Neither the European Union nor the granting authority can be held responsible for them.
        }%
    \thanks{
        $^{1}$Department of Computer, Control and Management Engineering, Sapienza University of Rome, 00185 Rome, Italy. {\tt\footnotesize schol@r-rapuano.com}, {\tt\footnotesize schol@r-franchi.eu}
        }%        
    \thanks{
        $^{2}$Robotics and Mechatronics Department, Electrical Engineering,  Mathematics, and Computer Science (EEMCS) Faculty, University of Twente, 7500 AE Enschede, The Netherlands.  {\tt\footnotesize y.shen-2@utwente.nl}, {\tt\footnotesize f.califano@utwente.nl}, {\tt\footnotesize c.gabellieri@utwente.nl}, {\tt\footnotesize schol@r-franchi.eu}.
        }%
    }
}
\ifwithauthors
    \myauthors
\fi

\begin{document}
    \maketitle
    \copyrightnotice
    \pagestyle{empty}

    \begin{abstract}
     This paper presents a framework for aerial manipulation of an extensible cable that combines a high-fidelity model based on partial differential equations (PDEs) with a reduced-order representation suitable for real-time control. The PDEs are discretised using a finite-difference method, and proper orthogonal decomposition is employed to extract a reduced-order model (ROM) that retains the dominant deformation modes while significantly reducing computational complexity. Based on this ROM, a nonlinear model predictive control scheme is formulated, capable of stabilizing cable oscillations and handling hybrid transitions such as payload attachment and detachment. Simulation results confirm the stability, efficiency, and robustness of the ROM, as well as the effectiveness of the controller in regulating cable dynamics under a range of operating conditions. Additional simulations illustrate the application of the ROM for trajectory planning in constrained environments, demonstrating the versatility of the proposed approach. Overall, the framework enables real-time, dynamics-aware control of unmanned aerial vehicles (UAVs) carrying suspended flexible cables.
\end{abstract}

    \section{INTRODUCTION}
\begin{figure}[t]
    \centering
    \includegraphics[width=0.6\linewidth]{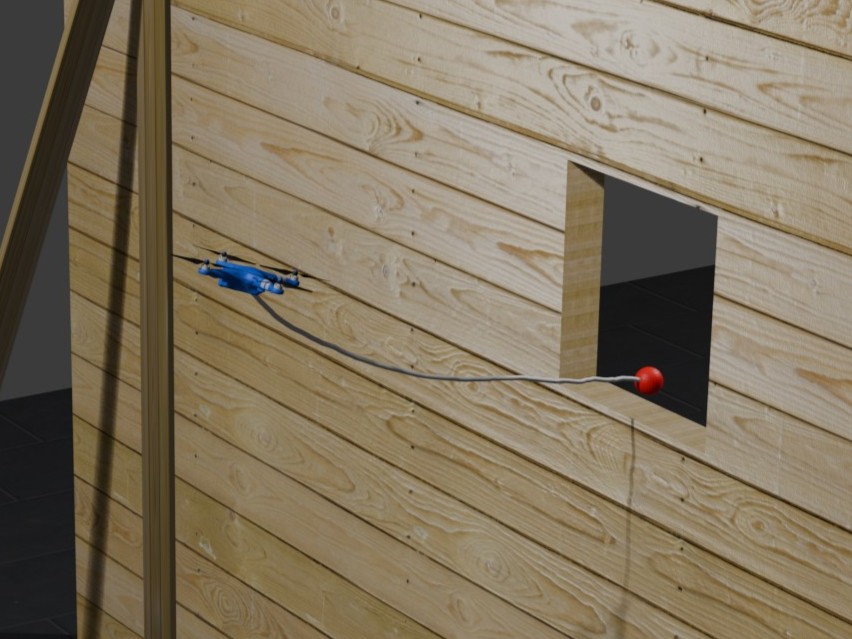}
    \caption{A UAV transports a slung payload through a narrow opening by exploiting the deformation of the cable.}
    \label{fig:hero}
\end{figure}

     Flexible cables offer a versatile alternative to rigid appendages in aerial manipulation. By keeping the vehicle at a distance, they enhance safety when interacting with hazardous or delicate objects, and their compliant nature makes them well-suited for operation in confined or cluttered environments. In addition, as illustrated in Fig.~\ref{fig:hero}, the passive dynamics of the cable may be leveraged to facilitate certain maneuvers, potentially reducing actuation effort.

    These advantages, however, come with challenges: cables are continuum systems governed by nonlinear dynamics, and when interacting with payloads, they exhibit hybrid behaviors due to switching boundary conditions at the tip. 

    \subsection{Related Work}
    Research on unmanned aerial vehicles (UAVs) with cable-suspended payloads can be broadly categorized by the level of fidelity in the cable model.
    \paragraph*{Rigid pendulum models} These represent the payload either as a single rigid body or as a point mass connected by a link. Such formulations enable efficient swing attenuation and are also used in cooperative transport \cite{2012-PalFieCru, 2017-FoeFalKupNavTedSca, 2023-GooBecCol, 2023-LiLoi}. Their main limitation is that they neglect slackness and distributed deformation, restricting applicability to scenarios where the cable remains taut and aligned beneath the vehicle.
  \paragraph*{Hybrid slack/taut models} In these approaches, the payload is modeled as switching between free-fall and tensioned regimes \cite{2013-SreLeeKum, 2017-KotWuSre, 2025-SarLiLoi}. This allows explicit handling of events such as re-tensioning, enabling controllers to maintain stability and performance through mode transitions. However, the underlying dynamics typically remain simplified.
 \paragraph*{Reduced-order continuum models} The authors of \cite{2018-GouDur} pioneered the usage of proper orthogonal decomposition (POD) to reduce the state dimension of soft robot models, allowing fast, but still meaningful, reduced-order models (ROMs). In fact, \cite{2025d-SheFraGab} applied POD to approximate the cable manipulated by a UAV by extracting its principal modes, enabling accurate capture of the deformation in a way suitable for model predictive control (MPC); However, \cite{2025d-SheFraGab} neglected any payload interactions.

    Each modeling choice embodies a trade-off between fidelity and tractability. Importantly, none of the existing approaches simultaneously combine accurate continuum modeling with hybrid payload dynamics in a form suitable for online control.
    
    \subsection{Contribution}
    This paper proposes a hybrid control framework for UAVs hosting flexible, extensible cables and executing dynamic payload manipulation. The main contributions are:
    \begin{itemize}[leftmargin=*]
        \item A hybrid model based on partial differential equations (PDEs) that represents the cable as a flexible continuum and explicitly captures payload attachment and detachment events.
        \item A POD-based reduced-order model (ROM) specifically adapted to preserve the dominant deformation modes and ensure modal coherence across hybrid transitions.
        \item A nonlinear MPC scheme that exploits the ROM to achieve real-time trajectory tracking while actively suppressing undesired cable dynamics, ensuring both accuracy and robustness during manipulation tasks.
    \end{itemize}
    Crucially, this framework is not just a refinement of existing approaches. Its novelty lies in the first integration of continuum tether modeling, model reduction, and hybrid predictive control into a single pipeline that is both theoretically sound and practically deployable. This combination provides tangible advantages in accuracy, robustness, and real-time feasibility, significantly extending the range of scenarios that can be addressed compared to prior methods.
    \section{BACKGROUND}

\subsection{Partial Differential Modeling of Extensible Strings}
\label{subsec:pde}
\begin{figure}[t]
    \centering
    \begin{tikzpicture}[scale = 1]
        % cable
        \draw plot [smooth, tension = 0.6, color = dark_gray] 
            coordinates {(0, 0) (0.2, -0.6) (1, -0.1) (1.3, -0.9) (2.4, -0.1) (2.6, -1.3) (3.3, -0.6) (3.5, -1.2) (4.2, -0.2)};
        \node [mark size = 1.5 pt] at (0, 0) {\pgfuseplotmark{*}};
        \node at (-0.6, 0.2) {$\bm{r}(0)$};
        \node [mark size = 1.5 pt] at (2.4, -0.1) {\pgfuseplotmark{*}};
        \node at (2.0, 0.2) {$\bm{r}(s)$};
        \node [mark size = 1.5 pt] at (4.2, -0.2) {\pgfuseplotmark{*}};
        \node at (4.8, 0) {$\bm{r}(L)$};
        \draw[->, thick, orange] (2.4, -0.1) -- ++(0.8, -0.25) node[anchor=south] {$\bm{r}_s(s)$};

        % world frame
        \draw[very thick, ->, color = orange, text = black] (- 2.5, - 2.5, - 2.5) -- ({- 2.5 + 1}, {- 2.5}, {- 2.5}) node[right] {$\bm{E}_x$};
        \draw[very thick, ->, color = blue, text = black] (- 2.5, - 2.5, - 2.5) -- ({- 2.5},{- 2.5 + 1},{- 2.5}) node[right] {$\bm{E}_z$};
        \draw[very thick, ->, color = green, text = black] (- 2.5,- 2.5, - 2.5) -- (- 2.5, - 2.5, - 3.5) node[right] {$\bm{E}_y$};
        \node[draw=none] at (- 2.75, - 2.6, - 2.5) {$\bm{O}$};
    \end{tikzpicture}
    \caption{Extensible string in $\mathbb{R}^3$.}
    \label{fig:string}
\end{figure}
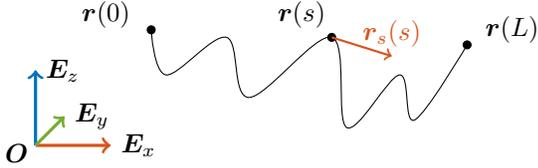

    Consider a perfectly flexible string of length $L$, parametrized by the curvilinear coordinate $s \in [0, L]$ and time $t$. In the world frame $\mathcal{F}?w = \{\bm{O}, \bm{E}_x, \bm{E}_y, \bm{E}_z\}$, its configuration is given by $\bm{r}(s, t) \in \mathbb{R}^3$ (see Fig.~\ref{fig:string}). Using Lagrange's notation, $\bm{r}_t \coloneq \frac{\partial\bm{r}}{\partial t}$ and $\bm{r}_{tt} \coloneq \frac{\partial^2 \bm{r}}{\partial t^2}$ denote velocity and acceleration, and $\bm{r}_s \coloneq \frac{\partial \bm r}{\partial s}$ the tangent vector.
    For compactness, the explicit time dependence $(t)$ is omitted throughout this paper unless required for clarity.
    
    The internal contact force at $s$, i.e.\ $\bm{n}(s)$, represents the force transmitted through the cross-section of the string, exerted by the portion $(s,L)$ on the portion $(0,s)$. Because of the assumption of perfect flexibility, the constitutive equation links the string deformation to the contact force through
    \begin{equation}
        \bm{n}(s) = E A \left(1 - \frac{1}{\|\bm{r}_s(s)\|}\right) \bm{r}_s(s), \label{eq:n}
    \end{equation}
    with $\|\cdot\|$ denoting the $L^2$-norm, $E$ being the Young modulus of the string, and $A$ its cross-sectional area.
    
    Take a string material segment $(s_0, s_1)$, with $0 < s_0 < s_1 < L$, subject to the contact forces $-\bm{n}(s_0)$ and $\bm{n}(s_1)$ and to the body force per unit length $\bm{\beta}(s)$.
    Below, the integral equation expresses the conservation of linear momentum for the material segment:
    \begin{multline}
        \frac{\mathrm{d}}{\mathrm{d}t} \int_{s_0}^{s_1} \rho?c A \bm{r}_t(s) \mathrm{d} s = \int_{s_0}^{s_1} \bm{\beta}(s) \mathrm{d} s - \bm{n}(s_0) + \bm{n}(s_1), \\
        \forall (s_0, s_1) \subset (0, L),
        \label{eq:integral}
    \end{multline}
    where $\rho?c$ is the density of the string. Differentiation returns the classical form:
    \begin{equation}
        \rho?c A \bm{r}_{tt}(s) = \bm{\beta}(s) + \bm{n}_s(s), \quad \forall s \in (0, L), \label{eq:classical}
    \end{equation}
    where $\bm{n}_s(s) \coloneq \frac{\partial \bm{n}}{\partial s} (s)$.
    At the endpoints, boundary conditions model the interaction of the string with any attached bodies.

    A  finite-difference method (FDM) is used to solve the string PDEs \cite{2007-GroRooSty}. The idea is to discretize the domain $[0, L]$ in $N$ intervals of length $h?s \coloneq \frac{L}{N}$, approximating the continuum as a chain of $N+1$ \textit{nodes}. For any field quantity, for instance $\bm{r}(s)$, its value at node $i$ is denoted $\bm{r}^i \coloneq \bm{r}(i \, h?s)$. Nodal velocities, accelerations and contact forces are written $\bm{r}_t^i$, $\bm{r}_{tt}^i$ and $\bm{n}^i$, respectively.
    
    The spatial derivatives are approximated using the central differencing scheme. For instance
    \begin{equation}
        \label{eq:r_der}
        \begin{gathered}
            \bm{r}_s^i \approx \frac{\bm{r}^{i+1} - \bm{r}^{i-1}}{2 h?s}, \quad i = 1, \dots, N - 1.
        \end{gathered} 
    \end{equation}
    Accordingly, the term $\bm{n}_s(s)$ becomes
    \begin{multline}
            \bm{n}_s^i \approx EA \Bigg( \frac{\bm{r}^{i+1} - 2 \, \bm{r}^i + \bm{r}^{i-1}}{h?s^2} + \frac{1}{h?s} \frac{\bm{r}^i - \bm{r}^{i-1}}{\| \bm{r}^i - \bm{r}^{i-1} \|} \\ 
            - \frac{1}{h?s} \frac{\bm{r}^{i+1} - \bm{r}^i}{\| \bm{r}^{i+1} - \bm{r}^i \|}\Bigg), \quad i = 1, \dots, N - 1. 
            \label{eq:n_der}
    \end{multline}
    Targeting \eqref{eq:classical}, the result is a system of $3(N+1)$ second-order ordinary differential equations. On the interior of the domain, the discretized version of the equations of motion reads
    \begin{equation}
        \rho?c A \bm{r}^i_{tt} = \bm{\beta}^i + \bm{n}_s^i, \quad i = 1, \ldots, N - 1.
        \label{eq:classical_fdm}
    \end{equation}
    At the boundaries, on the other hand, the equations depend on the specific boundary conditions imposed. 

    Spatial integrals can be approximated using the trapezoidal rule. In fact, \eqref{eq:integral} is approximated as
    \begin{multline}
        \frac{\rho?c A h?s}{2} \sum_{i = i_0 + 1}^{i_1} \left(\bm{r}_{tt}^i + \bm{r}_{tt}^{i - 1}\right) = \frac{h?s}{2} \sum_{i = i_0 + 1}^{i_1} \left(\bm{\beta}^i + \bm{\beta}^{i -1}\right) \\ 
        - \bm{n}^{i_0} + \bm{n}^{i_1}, \quad \forall i_0, i_1 \in \mathbb{N} \text{ s.t. } 0 \leq i_0 < i_1 \leq N.
        \label{eq:integral_fdm}
    \end{multline}

    To learn more about this section, the reader is referred to \cite{2005-Ant, 2025d-SheFraGab}.

\subsection{Model Order Reduction via Proper Orthogonal Decomposition}
    \label{sec:reduction}
    POD is a standard technique for reducing the complexity of a physical field by extracting a finite number of its most significant modes \cite{2017-TaiBruDawRowColMckSchGorTheUke}. A key step in POD is to project the target function onto a reduced-order functional basis \cite{2018-GouDur, 2017-TaiBruDawRowColMckSchGorTheUke}. For this purpose, one can decouple time and space by combining purely-spatial orthonormal modes via time-dependent coefficients. For a real-valued vector function $\bm{x}(s, t) \in \mathbb{R}^n$ (with $n$ the dimension of the embedding space, $n=3$ in the three-dimensional Euclidean space) defined over a one-dimensional domain $\mathcal{D}$, one has
    \begin{equation}
        \bm{x}(s, t) \approx \sum_{m} \bm{a}_m(t) \phi_m(s).
        \label{eq:pod_approx_back}
    \end{equation}
    Here, $\bm{a}_m(t) \in \mathbb{R}^n$ and $\phi_m(s) \in \mathbb{R}$ represent the $m$-th time-dependent coefficient and spatial basis function, respectively.

    $O + 1$ snapshots of the target function are collected over $M + 1$ spatial points into a tensor $\bm{\mathcal{X}} \in \mathbb{R}^{n \times (M + 1) \times (O + 1)}$. Each page of the tensor is a snapshot:
    \begin{multline}
        \bm{\mathcal{X}}_{::j} \coloneq \begin{bmatrix}
            \bm{x}(s_1, t_j) & \bm{x}(s_2, t_j) & \dots & \bm{x}(s_{M + 1}, t_j)
        \end{bmatrix}, \\
        j = 1, \dots, O + 1,
    \end{multline}
    with $s_j \in \mathcal{D}, \forall j = 1, \dots, M + 1$ being the sampling points. Singular value decomposition (SVD) of the mode-$2$ unfolding of $\bm{\mathcal{X}}$ breaks it down as $\bm{\mathbf{X}}_{(2)} = \mathbf{U}_2 \mathbf{\Sigma}_2 \mathbf{V}_2^\top.$ The spatial modes are given by the columns of the unitary matrix $\mathbf{U}_2 \in \mathbb{R}^{(M + 1) \times (M + 1)}$: 
    \begin{equation}
        \mathbf{U}_2 = \begin{bmatrix}
            \bm{u}_1 & \bm{u}_2 & \dots & \bm{u}_m & \dots & \bm{u}_{M + 1}
        \end{bmatrix}.
    \end{equation}
    In particular, their samples are obtained by normalization. With uniform spatial grid, one has
    \begin{multline}
        \bm{\phi}_m = \begin{bmatrix} \phi_m(s_1) & \phi_m(s_2) & \dots & \phi_m(s_{M + 1}) \end{bmatrix}^\top \coloneq \varphi \bm{u}_m, \\ 
        m = 1, \dots, M+1.
        \label{eq:pod_modes}
    \end{multline}
    The normalization factor $\varphi \in \mathbb{R}$ ensures orthonormality of the continuous modes over $\mathcal{D}$, i.e.
    \begin{equation}
        \delta_{jk} = \int_\mathcal{D} \phi_j(s) \phi_k(s) \, \mathrm{d}s, \quad j, \, k = 1, \dots, M + 1.
        \label{eq:orthonormality}
    \end{equation}

    Since at most $\min\{M + 1, n(O + 1)\}$ independent modes can be extracted, the number of snapshots should be chosen to satisfy $O + 1 \geq (M + 1)/n$; however, a larger $O$ increases representativeness. Hence, \eqref{eq:pod_approx_back} can be specified as
    \begin{equation}
        \begin{gathered}
            \bm{x}(s) \approx \sum_{m=1}^{M + 1} \bm{a}_m \phi_m(s).
        \end{gathered}
        \label{eq:pod_approx_back2}
    \end{equation}
    Vice versa, the coefficients $\bm{a}_m$ are found via the exploitation of the orthonormality condition \eqref{eq:orthonormality}:
    \begin{equation}
        \begin{gathered}
            \bm{a}_m = \int_\mathcal{D} \bm{x}(s) \phi_m(s) \, \mathrm{d}s.
        \end{gathered}
    \end{equation}
    The extension to the time derivatives, used to obtain the reduced-order dynamics, is straightforward, as one only needs to differentiate with respect to $t$.

    Further information about the application of POD to continuum mechanics can be found in \cite{2017-TaiBruDawRowColMckSchGorTheUke, 2018-GouDur, 2025d-SheFraGab}.
    \section{METHODOLOGY}

\subsection{Hybrid Model}
    \label{subsec:hybrid}
    To simplify the problem, the following choices are made:
    \begin{itemize}[leftmargin=*]
        \item The air is assumed to be homogeneous with constant properties, static, and all the aerodynamic disturbances are neglected except for a simplified drag force.
        \item The cable is modeled as an extensible string with perfect flexibility, with no internal damping nor self-interaction.
        \item Pick-and-place is abstracted as an instantaneous inelastic collision with an accompanying change of the tip mass, without explicitly modeling the attachment mechanism.
    \end{itemize}
    Every element of the cable is subject to gravitational and aerodynamic drag forces. The classical equation of motion \eqref{eq:classical_fdm} becomes
        \begin{equation}
            \rho?c A \bm{r}_{tt}^i = - \rho?c A g_0 \bm{E}_z - b?c \|\bm{r}_t^i\| \bm{r}_t^i + \bm{n}_s^i, \quad i = 1, \ldots, N - 1,
            \label{eq:complete}
        \end{equation}
    where $g_0 = \qty{9.81}{\meter \per \second \squared}$ is gravity and $b?c$ is the cable drag coefficient.

    To apply the central differencing scheme consistently at the boundary at $s = 0$, the contact force $\bm{n}^0$ is obtained from the approximated integral balance \eqref{eq:integral_fdm} with $i_0 = 0$ and $i_1 = 1$:
    \begin{multline}
        \bm{n}^0 = \bm{n}^1 - \rho?c A h?s g_0 \bm{E}_z \\
        - \frac{h?s}{2} [\rho?c A (\bm{r}_{tt}^0 + \bm{r}_{tt}^1) + b?c (\|\bm{r}_t^0\| \bm{r}_t^0 + \|\bm{r}_t^1\| \bm{r}_t^1)].
        \label{eq:n0}
    \end{multline}
    The boundary is attached to the UAV, modeled as a point mass $m?b$ subject to a control force $\bm{f}$ and gravity. Its interaction with the cable is described by
    \begin{equation}
        m?b \bm{r}_{tt}^0 = \bm{f} - m?b g_0 \bm{E}_z + \bm{n}^0. \label{eq:quadrotor} 
    \end{equation}
    Combining \eqref{eq:n0} and \eqref{eq:quadrotor} yields an explicit expression for the acceleration:
    \begin{multline}
        \bm{r}_{tt}^0 = \frac{1}{m?b + \frac{\rho?c A h?s}{2}} \bigg\{ \bm f - (m?b + \rho?c A h?s) g_0 \bm{E}_z \\
        - \frac{h?s}{2} \left[b?c \left(\|\bm{r}_t^0\| \bm{r}_t^0 + \|\bm{r}_t^1\| \bm{r}_t^1\right) + \rho?c A \bm{r}_{tt}^1\right] + \bm{n}^1 \bigg\}.
        \label{eq:r_tt_0}
    \end{multline}

    When nothing is attached to the distal end of the cable, \eqref{eq:classical_fdm} is extended to $i = N$, with the constraint that in that point the strain is null, i.e.\ $\|\bm{r}_s^N\| = 1 \implies \bm{n}^N = \bm{0}$ \cite{2025d-SheFraGab}.
    To enforce this condition, a \textit{ghost} node at $i = N+1$ is added:
    \begin{equation}
        \bm{r}^{N + 1} = \bm{r}^{N - 1} + 2 \, h?s \frac{\bm{r}^N - \bm{r}^{N - 1}}{\|\bm{r}^N - \bm{r}^{N - 1}\|}.
    \end{equation}
    
    Conversely, if a payload is attached to the tip, the unstrained condition is no longer valid. The boundary condition at $i = N$ is instead governed by the dynamics of the payload, which is described along the lines of \eqref{eq:quadrotor} with the addition of a drag term:
    \begin{equation}
        m?p \bm{r}_{tt}^N = - m?p g_0 \bm{E}_z - b?p \|\bm{r}_t^N\|\bm{r}_t^N -\bm{n}^N, \label{eq:payload}
    \end{equation}
    where $m?p$ is the mass of the payload and $b?p$ is its drag coefficient. A symmetric derivation with respect to \eqref{eq:r_tt_0} yields
    \begin{multline}
        \bm{r}_{tt}^N = \frac{1}{m?p + \frac{\rho?c A h?s}{2}} \bigg\{ - \bm{n}^{N-1} - \left(\frac{h?s b?c}{2} + b?p\right)\|\bm{r}_t^N\|\bm{r}_t^{N} - \\
        \frac{h?s}{2} \left( b?c \|\bm{r}_t^{N-1}\|\bm{r}_t^{N-1} + \rho?c A \bm{r}_{tt}^{N-1}\right) - (m?p + \rho?c A h?s) g_0 \bm{E}_z\bigg\}.
    \end{multline}

    Let $\bm{\nu}?p$ denote the payload velocity, and $t^-$, $t^+$ the instants immediately before and after payload pickup, respectively. 
    Transitions arise from two possible events:
    \begin{itemize}[leftmargin=*]
        \item \textit{Attachment}. 
        Once the payload attaches to the cable tip, its velocity must coincide with that of the distal node until detachment, i.e.\ $
            \bm{\nu}?p(t) = \bm{r}_t^N(t),$ $t \geq t^+$.
        At the instant of attachment, the inelastic collision between the payload and the tip mass lump $\rho?c A h?s/2$ enforces conservation of linear momentum:
        \begin{equation}
            m?p \bm{\nu}?p(t^-) + \frac{\rho?c A h?s}{2} \bm{r}_t^N(t^-) 
            = \left(m?p + \frac{\rho?c A h?s}{2}\right) \bm{r}_t^N(t^+).
        \end{equation}
        The resulting velocity reset at node $N$ is
        \begin{equation}
            \bm{r}_t^N(t^+) =
            \frac{1}{1 + \frac{\rho?c A h?s}{2 m?p}} \bm{\nu}?p(t^-) 
            + \frac{1}{1 + \frac{2 m?p}{\rho?c A h?s}} \bm{r}_t^N(t^-).
            \label{eq:ha_attach}
        \end{equation}

        \item \textit{Detachment}. 
        When the payload is released, the continuous state of the system is left unchanged. The two subsystems then evolve independently under their respective dynamics.
\end{itemize}

\subsection{POD-Based Model Order Reduction}
    \label{subsec:rom}
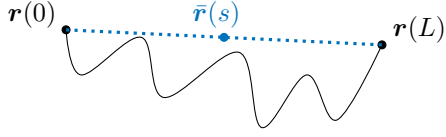
\begin{figure}[t]
    \centering
    \begin{tikzpicture}[scale = 1]
        % cable
        \draw plot [smooth, tension = 0.6, color = dark_gray] 
            coordinates {(0, 0) (0.2, -0.6) (1, -0.1) (1.3, -0.9) (2.3, -0.3) (2.6, -1.3) (3.2, -0.6) (3.6, -1.2) (4.2, -0.2)};
        \node [mark size = 1.5 pt] at (0, 0) {\pgfuseplotmark{*}};
        \node at (-0.45, 0.2) {$\bm{r}(0)$};
        \node [mark size = 1.5 pt] at (4.2, -0.2) {\pgfuseplotmark{*}};
        \node at (4.7, 0) {$\bm{r}(L)$};

        % segment
        \draw[line width = 1.2, dotted, color = blue] (0, 0) -- (4.2, -0.2);
        \node [mark size = 1.5 pt, color = blue] at (2.1, - 0.1) {\pgfuseplotmark{*}};
        \node [text = blue] at (2, 0.2) {$\bar{\bm{r}}(s)$};
    \end{tikzpicture}
    \caption{Decomposition of the string position map on which the proposed model-order reduction method is based.}
    \label{fig:rom}
\end{figure}

    Given only the solution at the boundaries (i.e.\ $\bm{r}^0$ and $\bm{r}^N$), the most natural choice is to place the intermediate nodes along the straight segment joining the two extremities (see Fig.~ \ref{fig:rom}). This reference configuration $\bar{\bm{r}}$ is defined as
    \begin{equation}
        \bar{\bm{r}}^i = \bm{r}^0 + \frac{i}{N} \, (\bm{r}^N - \bm{r}^0), 
        \quad i = 0, \dots, N. 
        \label{eq:segment}
    \end{equation}
    The actual node positions are then decomposed into the reference configuration plus a fluctuation term:
    \begin{equation}
        \bm{r}^i = \bar{\bm{r}}^i + \tilde{\bm{r}}^i, 
        \quad \tilde{\bm{r}}^i \coloneq \bm{r}^i - \bar{\bm{r}}^i. 
        \label{eq:decomposition}
    \end{equation}
    By construction, the fluctuation field satisfies homogeneous boundary condition $\tilde{\bm{r}}^0 = \tilde{\bm{r}}^N = \bm{0}$.

    One can construct a tensor of snapshots $\tilde{\bm{\mathcal{R}}}$ such that
        \begin{multline}
            \tilde{\bm{{\mathcal{R}}}}_{::j} = \begin{bmatrix}
                \tilde{\bm{r}}^0(t_j) & \tilde{\bm{r}}^d(t_j) & \tilde{\bm{r}}^{2d}(t_j) & \dots & \tilde{\bm{r}}^N(t_j)
            \end{bmatrix}, \\
            \quad j = 0, \dots, O,
        \end{multline}
    with $d \coloneq N/M \in \mathbb{N}$ being a uniform spatial decimation factor. Following the POD procedure, SVD returns singular vectors $\bm{u}_m$ with null first and last components. The actual POD modes are then obtained via \eqref{eq:pod_modes} after normalization. The trapezoidal approximation of the orthonormality condition \eqref{eq:orthonormality} with step size $h?d = h?s \, d$ reads
    \begin{equation}
    h?d \, \bm{\phi}_j^\top \bm{\phi}_k 
    = h?d \varphi^2 \bm{u}_j^\top \bm{u}_k 
    = h?d \varphi^2 \delta_{jk},
    \end{equation}
    which fixes the normalization factor as $\varphi = (1/h?d)^{1/2}$.

    The idea is to project the FDM model onto the space spanned by the subset of $R \leq M - 1$ orthonormal components whose energy, measured in terms of the magnitude of their corresponding singular values, is the largest. Using the linear combination formula \eqref{eq:pod_approx_back2}, the fluctuating part of the node positions can be approximated as
    \begin{equation}
        \begin{bmatrix}
            \tilde{\bm{r}}^0(t) \\
            \tilde{\bm{r}}^d(t) \\
            \vdots \\
            \tilde{\bm{r}}^N(t)
        \end{bmatrix} = \bm{\Phi}
        \begin{bmatrix}
            \bm{a}_1(t) \\
            \bm{a}_2(t) \\
            \vdots \\
            \bm{a}_{R}(t)
        \end{bmatrix},  \quad 
        \begingroup
        \setlength
        \arraycolsep{2pt}
        \bm{\Phi} \coloneq \begin{bmatrix}
            \bm{\phi}_1 & \bm{\phi}_2 & \dots & \bm{\phi}_{R}
        \end{bmatrix} \otimes \mathbf{I}_3,
        \endgroup
        \label{eq:pod_approx}
    \end{equation}
    where $\otimes$ denotes the Kronecker product and $\mathbf{I}_n$ is the $n \times n$ identity matrix.
    
    Conversely, the projection onto the reduced subspace is performed using the Moore-Penrose pseudoinverse $\bm{\Phi}^+$ of $\bm{\Phi}$:
    \begin{equation}
        \begin{bmatrix}
            \bm{a}_1(t) \\
            \bm{a}_2(t) \\
            \vdots \\
            \bm{a}_{R}(t)
        \end{bmatrix} = \bm{\Phi}^+
        \begin{bmatrix}
            \tilde{\bm{r}}^0(t) \\
            \tilde{\bm{r}}^d(t) \\
            \vdots \\
            \tilde{\bm{r}}^N(t)
        \end{bmatrix}.
        \label{eq:pod_proj}
    \end{equation}
    
    The extension to the time derivatives is straightforward. Once the nodal accelerations $\bm{r}_{tt}^i$, $i = 0, d, \dots, N$, are computed from the equations of motion (Sec.\ \ref{subsec:hybrid}), they are projected via \eqref{eq:decomposition} and \eqref{eq:pod_proj} to obtain $\ddot{\bm{a}}$. Integration then yields $\dot{\bm{a}}$ and $\bm{a}$, which reconstruct the nodal states through \eqref{eq:pod_approx} and \eqref{eq:decomposition}.

    If an alternative basis $\bm{\Phi}^\prime$ is employed, the corresponding modal coordinates $\bm{a}^\prime$ relate to the previous ones by the transformation
    \begin{equation}
        \bm{a} = \bm{\Phi}^+ \bm{\Phi}^\prime \bm{a}^\prime.
        \label{eq:coord}
    \end{equation}
    Such mappings are useful, for instance, when switching between ROMs built from different training datasets.

    In conclusion, the method reduces the original FDM state space $\mathbb{R}^{6(N+1)}$ to $\mathbb{R}^{6(R+2)}$ --- accounting for two boundary points and $R$ modal coefficients. The larger $R$, the more accurate the approximation, at the cost of a higher state dimension.

    \subsection{Hybrid Nonlinear MPC}
    \label{subsec:hnmpc}

    Consider a discrete-time horizon of length $H$ indexed by $k \in \mathbb{N}$. The hybrid reduced-order internal model of the controller works with:
    \begin{itemize}
        \item a discrete state $q(k) \in \{0,1\}$, indicating whether the payload is detached or attached, respectively;
        \item a continuous state vector $\bm{z}(k) \in \mathbb{R}^{6(R+2)}$, collecting boundary positions and velocities together with the $R$ modal coefficients and their derivatives;
        \item the control input $\bm{v}(k) \in \mathbb{R}^3$, representing the UAV acceleration.
    \end{itemize}
    At the initial step, $q(0)$ and $\bm{z}(0)$ are set according to the current physical configuration, and in particular
    \begin{equation}
        \bm{z}^\top(0) \gets
        \begin{bmatrix}
            \bm{r}^{0\top} & \bm{a}_1^\top & \dots & \bm{a}_{R}^\top & \bm{r}^{N\top} &
            \bm{r}_t^{0\top} & \dot{\bm{a}}_1^\top & \dots & \dot{\bm{a}}_{R}^\top & \bm{r}_t^{N\top}
        \end{bmatrix}^\top.
        \label{eq:initial_state}
    \end{equation}

    The internal model of the controller is a ROM-based hybrid automaton, whose discrete successor is
    \begin{equation}
        (q(k+1), \bm{z}(k+1)) = \bm{T}\big(q(k), \bm{z}(k), \bm{v}(k)\big).
        \label{eq:internal_succ}
    \end{equation}
    $\bm{T}$ encapsulates both the (integrated) continuous reduced dynamics and the logic governing mode transitions. To ensure accuracy in both operating modes, two reduced bases are precomputed: $\bm{\Phi^0}$ for the free-tip configuration and $\bm{\Phi^1}$ for the slung-payload configuration. When no hybrid transition occurs, $\bm{z}(k)$ evolves by integrating the reduced dynamics of Sec.\ \ref{subsec:rom} --- based on the current discrete state $q(k)$ --- using $\bm{v}(k)$ in place of $\bm{r}_{tt}^0$. Conversely, if $\bm{z}(k+1)$ triggers a transition, in addition to updating $q(k+1)$:
    \begin{enumerate}[leftmargin=*]
        \item The modal coordinates (and their derivatives) within $\bm{z}(k+1)$ are transferred between bases using \eqref{eq:coord}.
        \item In the case of attachment, the tip velocity is updated according to the inelastic impact law \eqref{eq:ha_attach}.
    \end{enumerate}

    The hybrid NMPC is formulated as
    \begin{multline}
        \min_{\{\bm v(k)\}_{k=0}^{H-1}} 
        \sum_{k=0}^{H-1} \ell^{q(k), \hat q(k)}(\bm z(k), \hat{\bm{z}}(k), \bm v(k), \hat{\bm{v}}(k))
        \\
        + \ell?H^{q(k), \hat q(k)}(\bm z(H), \hat{\bm{z}}(H))
    \end{multline}
    \begin{equation}
            \text{s.t.} \quad
                \eqref{eq:initial_state} \quad \text{and} \quad
                \eqref{eq:internal_succ}, \quad k=0,\dots,H-1 ,
        \label{eq:mpc}
    \end{equation}
    where the hat symbol $\hat{}$ denotes reference state and input trajectories provided by an external planner. The running and terminal costs --- respectively, $\ell^{q, \hat q}$ and $\ell?H^{q, \hat q}$ --- penalize state and input tracking errors with respect to the references.
    
    To ensure consistency across hybrid transitions, the reduced states $\bm{z}, \bm{\hat{z}}$ are first mapped to the FDM space by linear projection matrices $\mathbf{P}^q , \mathbf{P}^{\hat{q}} \in \mathbb{R}^{6(N+1)\times 6(R+2)}$, constructed from the POD basis of the corresponding mode $q$:
    \begin{equation}
        \begin{gathered}
            \ell^{q, \hat q}(\bm z, \hat{\bm{z}}, \bm v, \hat{\bm{v}}) 
            = \| \mathbf{P}^q \bm z - \mathbf{P}^{\hat q} \hat{\bm{z}} \|_{\mathbf{S}}^2 
            + \| \bm v - \hat{\bm{v}} \|_{\mathbf W}^2,\\
            \ell?H^{q, \hat q}(\bm z, \hat{\bm{z}}) 
            = \| \mathbf{P}^q \bm z - \mathbf{P}^{\hat q} \hat{\bm{z}} \|_{\mathbf{S}?H}^2.
        \end{gathered}
        \label{eq:cost_node}
    \end{equation}
    where $\| \bm{x} \|_{\mathbf S}^2 \coloneq \bm{x}^\top \mathbf{S}\bm{x}$, and with $\mathbf S, \mathbf{S}?H \geq 0$ and $\mathbf W > 0$ being symmetric weighting matrices. 
    
    This construction yields a cost metric that is consistent across mode switches, since the error is always evaluated at the node level.
    Besides, soft state constraints can be incorporated by adding smooth barrier terms to $\ell^{q, \hat q}$ and $\ell?H^{q, \hat q}$.
    The optimization returns a sequence of predicted UAV acceleration commands, of which the first is applied to the vehicle in receding-horizon fashion. To attenuate discontinuities between successive MPC calls, the actual control signal is generated by interpolating between the first two predicted commands. Hybrid iLQR \cite{2021-KonCouJoh} is adopted as a solver due to its efficiency and its ability to handle discrete transitions, although other nonlinear programming methods could be employed.

   \color{black}
    \section{NUMERICAL EXPERIMENTS}
    The hybrid model of Sec.\ \ref{subsec:hybrid} is simulated with $N = 100$ using a fourth-order Runge--Kutta scheme with step size $\qty{5e-4}{\second}$ ($\qty{2}{\kilo\hertz}$). The physical and geometrical parameters are listed in Tab.\ \ref{tab:param}. The UAV employed in this study is a quadrotor (see Fig.~\ref{fig:system}), stabilized by a geometric controller \cite{2010-LeeLeoMcc} running every $\qty{5e-3}{\second}$ ($\qty{200}{\hertz}$). The MPC runs with sampling time $\qty{2.5e-2}{\second}$ ($\qty{40}{\hertz}$) and prediction horizon $H = 32$ --- implying that it predicts $\qty{0.8}{\second}$ into the future. This suffices to capture the onset of the dominant swing modes while limiting the computational footprint.

    \begin{figure}[t]
    \centering
    \tdplotsetmaincoords{65}{10}

    \begin{tikzpicture}[tdplot_main_coords, scale=1]
        % quadrotor
        \draw[line width = 2, color = gray] (-1, 0, 1) -- (1, 0, 1);
        \draw[line width = 2, color = gray] (0, -1, 1) -- (0, 1, 1);

        % rotors
        \draw[draw = none, fill = gray!40, opacity = 0.6] (0, 1, 1) circle (0.5);
        \draw[draw = none, fill = gray!40, opacity = 0.6] (1, 0, 1) circle (0.5);
        \draw[draw = none, fill = gray!40, opacity = 0.6] (0, - 1, 1) circle (0.5);
        \draw[draw = none, fill = gray!40, opacity = 0.6] (- 1, 0, 1) circle (0.5);

        % cable
        \draw plot [smooth, tension = 0.6, color = dark_gray] coordinates {(0, 0, 1) (0.4, 0, 0.5) (0.1, 0, 0) (0.5, 0, - 0.5) (0.2, 0, - 1) (0.6, 0, - 1.5) (0.2, 0, - 2)};
        \node [mark size = 1.5 pt] at (0.5, 0, -0.5) {\pgfuseplotmark{*}};
        \node at (0.95, 0.2, -0.5) {$\bm{r}(s)$};
        \node [mark size = 1.5 pt] at (0, 0, 1) {\pgfuseplotmark{*}};
        \node at (- 0.4, - 0.4, 1) {$\bm{r}(0)$};

        % payload
        \node [mark size = 4 pt, color = orange] at (0.2, 0, - 2) {\pgfuseplotmark{*}};
        \node at (0.8, -0.1, -2) {$\bm{r}(L)$};	

        % world frame
        \draw[very thick, ->, color = orange, text = black] (- 2, - 2, - 1) -- ({- 2 + 0.8192}, {- 2 - 0.5736}, {- 1}) node[right] {$\bm{E}_x$};
        \draw[very thick, ->, color = green, text = black] (- 2, - 2, - 1) -- ({- 2 + 0.5736},{- 2 + 0.8192},{- 1}) node[right] {$\bm{E}_y$};
        \draw[very thick, ->, color = blue, text = black] (- 2,- 2, - 1) -- (- 2, - 2, 0) node[right] {$\bm{E}_z$};
        \node[draw=none] at (- 2.2, - 2.2, - 1) {$\bm{O}$};

        % gravity
        \draw[very thick, ->, color = black, text = black] (2.5, - 1, 0) -- node[right] {$- g_0 \bm{E}_z$} (2.5, -1, -1);
    \end{tikzpicture}
    \caption{Aerial system composed of a quadrotor with a flexible cable and a payload attached to the distal end.}
\label{fig:system}
\end{figure}
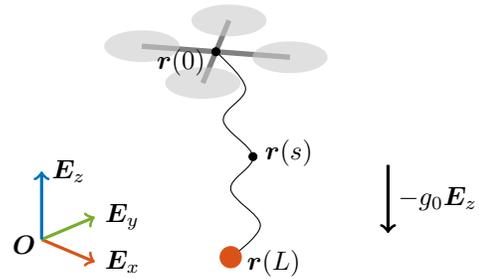
    \begin{table}[b]
    \caption{Physical and geometrical parameters of the aerial system.}
    \label{tab:param}
    \begin{center}
        \begin{tabular}{|c|c|c|c|}
            \hline
            \textbf{Symbol} & \textbf{Value [SI units]} & \textbf{Symbol} & \textbf{Value [SI units]} \\
            \hline
            $m?b$ & $0.3$ & $L$ & $1$ \\
            $m?p$ & $0.1$ & $\rho?c$ & $\qty{1.27e3}{}$ \\
            $b?p$ & $\qty{1.29e-2}{}$ & $A$ & $\qty{7.85e-5}{}$ \\  
            $b?c$ & $\qty{1.29e-2}{}$ & $E$ & $\qty{1e5}{}$ \\
            \hline
        \end{tabular}
    \end{center}
\end{table}

    Two solver variants are considered for the MPC, both implemented in CasADi \cite{2019-AndGilHorRawDie} for symbolic efficiency (without C code generation, leaving some margin for further speedup). The full \textit{HiLQR} incorporates numerical stabilization techniques such as line search and Levenberg--Marquardt regularization, whereas the simplified \textit{HiLQR-RTI} (\textit{real-time iteration}) performs only a single rollout, backward sweep, and forward sweep per iteration with regularization disabled.

    All simulations were performed in \textit{MATLAB R2024b} on a laptop with an \textit{Intel Core i5-1235U} CPU and $\qty{16}{\giga\byte}$ RAM.

\subsection{ROM Training and Evaluation}
    The POD basis was extracted from snapshots of the cable dynamics obtained during a $\qty{10}{\second}$ simulation starting from rest. To excite the dominant modes, the quadrotor was commanded to follow an isotropic sinusoidal trajectory designed to stimulate representative cable motions. This ensures the POD basis is robust and not overfitted to specific maneuvers. A spatial decimation factor $d = 10$ was adopted, yielding a sampling step $h?d = \qty{0.1}{\meter}$ ($M = 10$), and for the temporal snapshots $O = 50$. Separate POD bases were computed for the free-tip and slung-payload configurations.
    
\begin{figure}[t]
    \begin{subfigure}{0.49\linewidth}
        \centering
        \includegraphics[width=\linewidth]{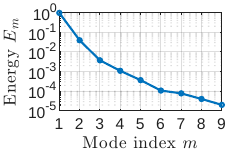}
        \caption{Free tip.} \label{fig:energy_free}
    \end{subfigure}
    \hspace*{\fill}
    \begin{subfigure}{0.49\linewidth}
        \centering
        \includegraphics[width=\linewidth]{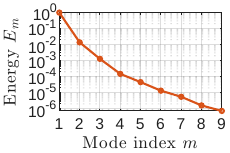}
        \caption{Slung payload.} \label{fig:energy_slung}
    \end{subfigure}

    \caption{
        Relative energy content of each independent mode found via POD (logarithmic scale).  
    }
    \label{fig:pod_energy}
\end{figure}

\begin{figure}[t]
    \begin{subfigure}{0.49\linewidth}
        \centering
        \includegraphics[width=\linewidth]{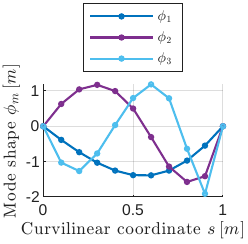}
        \caption{Free tip ($q = 0$).} \label{fig:modes_free}
    \end{subfigure}
    \hspace*{\fill}
    \begin{subfigure}{0.49\linewidth}
        \centering
        \includegraphics[width=\linewidth]{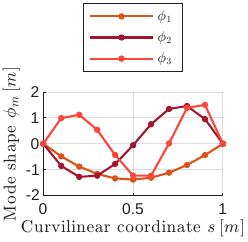}
        \caption{Slung payload ($q = 1$).} \label{fig:modes_slung}
    \end{subfigure}
    
    \caption{
        Spatial representation of the $3$ modes with the highest energy content, i.e.\ the largest singular values.  
    }
    \label{fig:pod_modes}
\end{figure}

    The relative energy contribution of mode $\bm{\phi}_m$ --- with singular value $\sigma_m$ --- is
    \begin{equation}
        E_m \coloneq \frac{\sigma_m^2}{\sum_{j=1}^{M-1} \sigma_j^2}, 
        \quad m = 1,\dots,M-1.
    \end{equation}
    As shown in Fig.~\ref{fig:pod_energy}, the first two modes capture more than $99\%$ of the energy in both cases, with mode $1$ alone exceeding $95\%$. The spatial structures of the modes resemble distorted sinusoids, especially near the free end in the free-tip case (Fig.~\ref{fig:pod_modes}).

    As a baseline for comparison, the POD-based ROM of \cite{2025d-SheFraGab} was adopted; the original formulation is designed for free-tip cables, and here it was adapted to include payload dynamics. Unlike the baseline, which projects all cable and payload states onto the reduced basis, the proposed method reduces only the dynamics of the interior while keeping the tip and payload equations intact, thus requiring one less reduced mode for the same state size. Both ROMs were then evaluated in a test case where the cable was initially released horizontally along the $x$-axis and the quadrotor was commanded to follow a $\qty{1}{\meter}$ translation along the $y$-axis under a quintic rest-to-rest timing law over $[0,1] \qty{}{\second}$. Under these conditions, the ROMs were integrated (with RK4) and compared against the full finite-difference model --- which serves as the high-fidelity ground truth --- assessing wall-clock integration time (step size $\qty{5e-4}{\second}$), maximum stable time step size, and approximation errors.

\begin{figure}[t]
    \begin{center}
        \begin{subfigure}{0.49\linewidth}
            \centering
            \includegraphics[width=\linewidth]{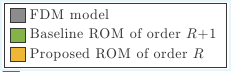}
        \end{subfigure}
    \end{center}
    \vspace*{-0.5em}
    \begin{subfigure}{0.49\linewidth}
        \centering
        \includegraphics[width=\linewidth]{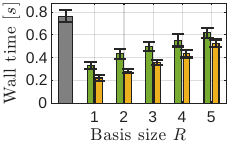}
        \caption{Statistics of the wall time required to integrate the models. Mean and standard deviation are determined from a population of $30$ observations per discrete state.} \label{fig:rom_wt}
    \end{subfigure}
    \hspace*{\fill}
    \begin{subfigure}{0.49\linewidth}
        \centering
        \includegraphics[width=\linewidth]{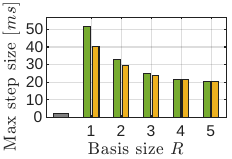}
        \caption{Maximum time step for which the models are stable. The minimum value encountered through trial-and-error among the two discrete states is selected.} \label{fig:rom_step}
    \end{subfigure}

    \caption{
        Time characteristics across model variants.
    }
    \label{fig:rom_time}
\end{figure}

    Results showed that, compared to the FDM-based model, both ROMs drastically reduced computation time, with the latter growing almost linearly with the number of retained modes $R$ (Fig.~\ref{fig:rom_wt}); the proposed approach was consistently faster due to the fewer projection operations required. Importantly, both ROMs admitted integration step sizes more than an order of magnitude larger than the FDM model (Fig.~\ref{fig:rom_step}), which further improves computational efficiency --- a critical feature for real-time applicability. 

\begin{figure}[t]
    \begin{subfigure}{0.49\linewidth}
        \centering
        \includegraphics[width=\linewidth]{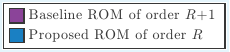}
    \end{subfigure}
    \hspace*{\fill}
    \begin{subfigure}{0.49\linewidth}
        \centering
        \includegraphics[width=\linewidth]{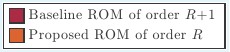}
    \end{subfigure}\\[1pt]
    \begin{subfigure}{0.49\linewidth}
        \centering
        \includegraphics[width=\linewidth]{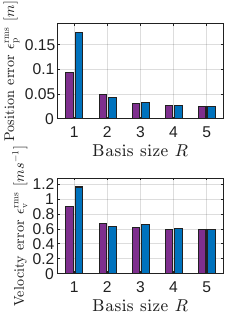}
        \caption{Free tip ($q = 0$).} \label{fig:rom_order_free}
    \end{subfigure}
    \hspace*{\fill}
    \begin{subfigure}{0.49\linewidth}
        \centering
        \includegraphics[width=\linewidth]{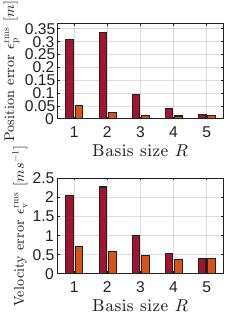}
        \caption{Slung payload ($q = 1$).} \label{fig:rom_order_slung}
    \end{subfigure}
    \caption{Accuracy of the ROMs with respect to the FDM model across different orders.}
    \label{fig:rom_order}
\end{figure}

    The approximation error between the ROM trajectories $(\bm{\mathfrak{r}}, \bm{\mathfrak{r}}_t)$ and the full FDM ones is measured with the functional
    \begin{equation}
            \epsilon(\bm{r}, \bm{\mathfrak{r}}) \coloneq \sqrt{\frac{h?d}{2L} \sum_{\substack{i=d \\ \text{step } d}}^N 
            \big(\| \bm{r}^i - \bm{\mathfrak{r}}^i\|^2 + \| \bm{r}^{i-1} - \bm{\mathfrak{r}}^{i-1}\|^2\big)},
            \label{eq:error}
    \end{equation}
    The root mean square value over the simulation horizon is denoted by the superscript $^{\mathrm{rms}}$. This definition is applied both to positions and velocities; in Fig.~\ref{fig:rom_order}, $\epsilon?p^{\mathrm{rms}} \coloneq \epsilon^{\mathrm{rms}}(\bm{r}, \bm{\mathfrak{r}})$ and $\epsilon?v^{\mathrm{rms}} \coloneq \epsilon^{\mathrm{rms}}(\bm{r}_t, \bm{\mathfrak{r}}_t)$. According to this metric, the accuracy was comparable in the free-tip case with at least from $R = 2$ (Fig.~\ref{fig:rom_order_free}), but with a payload attached the proposed ROM significantly outperformed the baseline, which was not tailored for payload dynamics (Fig.~\ref{fig:rom_order_slung}).

\begin{figure*}[t]
    \begin{center}
        \begin{subfigure}{\textwidth}
            \centering
            \includegraphics[width=\linewidth]{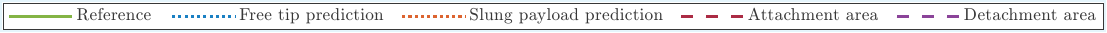}
            \end{subfigure}     
    \end{center}
    \begin{subfigure}{\textwidth}
        \centering
        \hspace*{\fill}
        \includegraphics[width=0.195\linewidth]{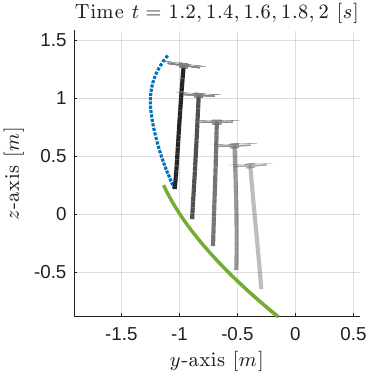}
        \hspace*{\fill} 
        \includegraphics[width=0.195\linewidth]{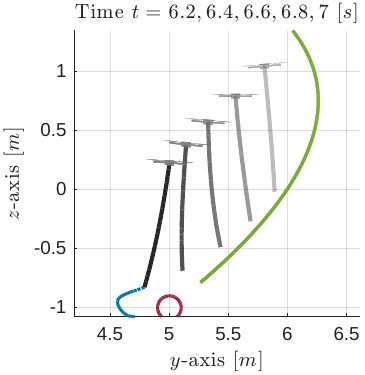}
        \hspace*{\fill}
        \centering
        \includegraphics[width=0.195\linewidth]{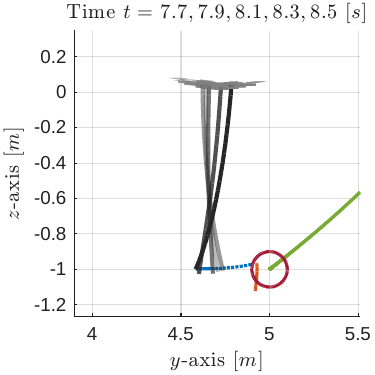}
        \hspace*{\fill}
        \centering
        \includegraphics[width=0.195\linewidth]{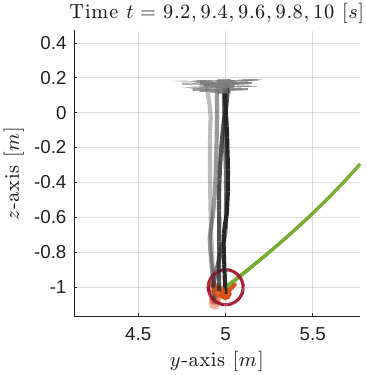}
        \hspace*{\fill}
        \caption{Trajectory tracking through a payload attachment event ($q = 0 \to 1$). Note the transition to the slung configuration. Solver: HiLQR-RTI.}
        \label{fig:track_motion_free}
    \end{subfigure}
    \\[15pt]
    \begin{subfigure}{\textwidth}
        \centering
        \hspace*{\fill}
        \includegraphics[width=0.195\linewidth]{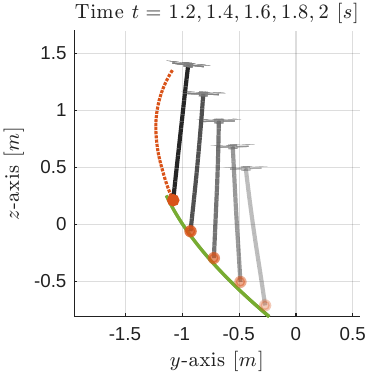}
        \hspace*{\fill}
        \centering
        \includegraphics[width=0.195\linewidth]{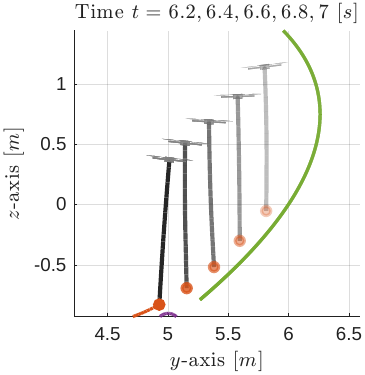}
        \hspace*{\fill}
        \centering
        \includegraphics[width=0.195\linewidth]{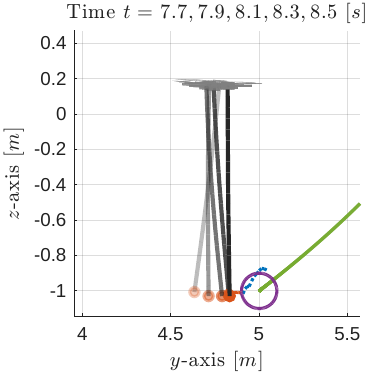}
        \hspace*{\fill}
        \centering
        \includegraphics[width=0.195\linewidth]{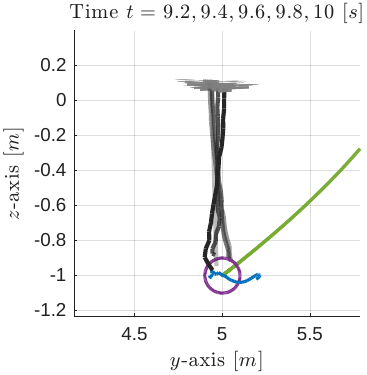}
        \hspace*{\fill}
        \caption{Trajectory tracking through a payload detachment event ($q = 1 \to 0$). The controller stabilizes the cable after release. Solver: HiLQR-RTI.}
        \label{fig:track_motion_slung}
    \end{subfigure}
    \\[15pt]
    \begin{subfigure}{\textwidth}
        \centering
        \includegraphics[width=0.195\linewidth]{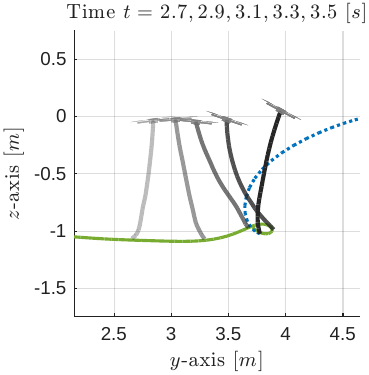}
        \centering
        \includegraphics[width=0.195\linewidth]{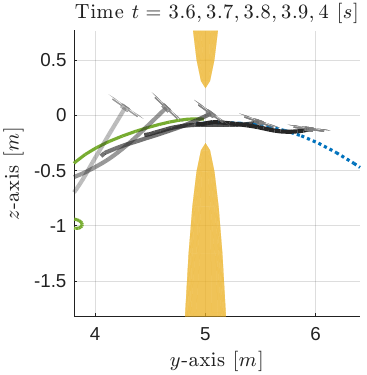}
        \centering
        \includegraphics[width=0.195\linewidth]{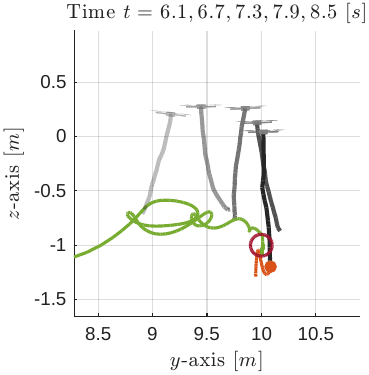}
        \centering
        \includegraphics[width=0.195\linewidth]{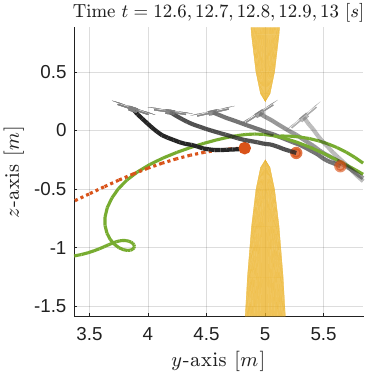}
        \centering
        \includegraphics[width=0.195\linewidth]{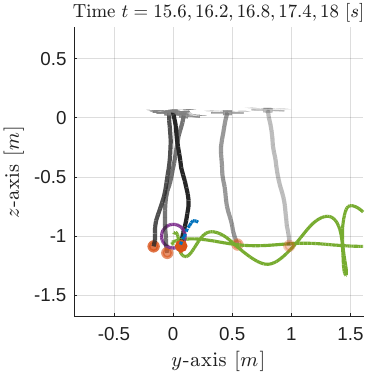}
        \caption{Rest-to-rest pick-and-place maneuver with obstacle avoidance (yellow ellipsoids). The planner generates a collision-free path through the window aperture, which is tracked by the controller to successfully complete the maneuver. Solver: HiLQR.}
        \label{fig:avoid_motion}
    \end{subfigure}
    
    \caption{Side-view time-lapses of the proposed Hybrid NMPC in experimental validation. The transparency indicates time evolution (opaque is current).}
    \label{fig:track_motion}
\end{figure*}
\subsection{Real-Time Manipulation}
    To showcase the control system, consider a scenario in which the aerial system must approach a distant point along a custom path and pick up/release the payload there. Starting from rest, the last node is commanded to track a spline trajectory $\left(\hat{\bm{r}}^N, \hat{\bm{r}}^N_t, \hat{\bm{r}}^N_{tt}\right)$ that interpolates preset waypoints with continuity up to the acceleration and quintic rest-to-rest timing law. The payload either starts at the last waypoint with zero velocity, or is already attached to the cable and must be released there.
    
    An internal ROM of order $R=1$ is sufficient to ensure satisfactory performance with minimal computational burden.

    Two $\qty{10}{\second}$ simulations were carried out, with and without the payload initially attached, as shown in Fig.~\ref{fig:track_motion_free} and \ref{fig:track_motion_slung}.
    In both cases, the cable and payload were successfully manipulated along the reference with limited error at the tip --- as reported in Tab.\ \ref{tab:track_res}. \textit{HiLQR-RTI} reduced the average MPC solution time $h?w$ to $\qty{16.4}{\milli\second}$ compared to $\qty{32.3}{\milli\second}$ for the full \textit{HiLQR}, confirming its suitability for real-time operation.
    The worst-case execution time of the RTI variant remained strictly below the $25$ ms sampling interval, preventing destabilizing overruns. The trade-off is that it is inherently less numerically robust, so the choice between the two solvers depends on the available computational resources and robustness requirements.

    \begin{table}[htbp]
    \caption{Cable-tip RMS tracking errors and average MPC iteration wall time in the online trajectory tracking scenario.}
    \label{tab:track_res}
    \begin{center}
    \begin{subtable}{\linewidth}
        \caption{Free-tip start.}
        \centering
        \begin{tabular}{|c|c|c|c|}
            \hline
            \textbf{Metric} & \textbf{HiLQR} & \textbf{HiLQR-RTI} & \textbf{Unit} \\
            \hline
            $\|\bm{r}^N - \hat{\bm{r}}^N\|^{\mathrm{rms}}$ & 0.194 & 0.174 & \qty{}{\meter} \\
            $\|\bm{r}_t^N - \hat{\bm{r}}_t^N\|^{\mathrm{rms}}$ & 0.449 & 0.369 & \qty{}{\meter \per \second}  \\
            $h?w$ & 32.3 & 16.4 & \qty{}{\milli \second} \\
            \hline
        \end{tabular}
    \end{subtable}
    \\[10pt]
    \begin{subtable}{\linewidth}
        \caption{Slung-payload start.}
        \centering
        \begin{tabular}{|c|c|c|c|}
            \hline
            \textbf{Metric} & \textbf{HiLQR} & \textbf{HiLQR-RTI} & \textbf{Unit} \\
            \hline
            $\|\bm{r}^N - \hat{\bm{r}}^N\|^{\mathrm{rms}}$ & 0.234 & 0.201 & \qty{}{\meter} \\
            $\|\bm{r}_t^N - \hat{\bm{r}}_t^N\|^{\mathrm{rms}}$ & 0.503 & 0.446 & \qty{}{\meter \per \second} \\
            $h?w$ & 36.5 & 17.8 & \qty{}{\milli \second} \\
            \hline
        \end{tabular}
    \end{subtable}
    \end{center}
\end{table}

\subsection{Planned Obstacle Avoidance}
    \label{subsec:avoid}

    A pick-and-place task with obstacle avoidance is the most challenging scenario addressed in this work. To this end, a custom segmented planner with log-barrier homotropy \cite{2021-TanChuJinAu} is employed as a pre-processing stage. The planner follows a divide-and-conquer approach in which the horizon is divided into shorter subproblems, optimized sequentially using the POD-based HiLQR (ROM with $R=2$ and step size $\qty{2.5e-3}{\second}$) and later stitched together into a globally feasible trajectory. A final refinement over the full horizon ensures smoothness, while the homotopy strategy gradually tightens the log-barrier terms.
    
    As shown in Fig.~\ref{fig:avoid_motion}, the task is to traverse a narrow window (at $y=\qty{5}{\meter}$, aperture $\qty{0.5}{\meter}$), retrieve a payload at $y=\qty{10}{\meter}$, and return --- all avoiding contact with the edges of the window. Attachment and detachment guards are defined around the pickup and drop-off locations, while the window edges are modeled as ellipsoidal exclusion regions infinite along the $x$-axis.
    
    Preset waypoints define the intermediate objectives of the maneuver. The planner uses them as attractors within each segment to shape the overall motion while respecting the constraints. Cooldown intervals of $\qty{2}{\second}$ with static references are inserted after each hybrid transition to allow the MPC to dissipate residual oscillations before proceeding. The weight matrices and barrier parameters are tuned to encourage waypoint tracking and collision avoidance.

    The results after $\qty{20}{\second}$ of simulation, summarized in Tab. \ref{tab:avoid_res}, confirm that the proposed strategy enables the UAV to accomplish the task without collisions. In this highly non-convex case, however, the simplifications of the HiLQR-RTI solver compromise robustness, and this variant fails to generate feasible solutions. The full HiLQR solver, while computationally heavier due to the additional stabilization steps, succeeds in completing the maneuver, highlighting the necessity of both the custom planning stage and the more stable solver for this level of complexity.

    \begin{table}[hbtp]
    \caption{RMS whole-cable tracking errors and average MPC iteration wall time in the obstacle avoidance experiment.
    }
    \label{tab:avoid_res}
    \begin{center}
        \begin{tabular}{|c|c|c|c|}
            \hline
            \textbf{Metric} & \textbf{HiLQR} & \textbf{HiLQR-RTI} & \textbf{Unit} \\
            \hline
            $\epsilon^{\mathrm{rms}}(\bm{r}, \hat{\bm{r}})$ & 0.153 & -- & \qty{}{\meter} \\
            $\epsilon^{\mathrm{rms}}(\bm{r}_t, \hat{\bm{r}}_t)$ & 0.941 & -- & \qty{}{\meter\per\second} \\
            $h?w$ & 42.8 & -- & \qty{}{\milli\second} \\
            \hline
        \end{tabular}
    \end{center}
\end{table}

    \section{CONCLUSIONS}
    This paper has presented a hybrid modeling and control framework for aerial payload manipulation with flexible cables, combining PDE-based continuum dynamics, POD-based model order reduction, and a nonlinear hybrid MPC scheme. The proposed reduced-order model achieved substantial computational savings while preserving accuracy, and its validation against both the full ground-truth model and a state-of-the-art ROM confirmed superior performance in scenarios involving payload dynamics. Embedding the ROM within a predictive control scheme enabled precise trajectory tracking in pick-and-place tasks, while its use within a custom planner paired with the controller ensured safe execution even in challenging obstacle-avoidance scenarios.

    The high-fidelity simulations use as ground truth a finite-difference model with fine spatial resolution and high integration frequency, whereas the controller operates on a much coarser reduced-order model. Despite this pronounced model–controller mismatch, the closed-loop performance highlights both the adequacy of the reduction strategy and the robustness of the proposed framework.
    
    Future work will address robustness analysis under uncertainties and disturbances, including the effects of internal damping and self-collision which may become significant during complex entanglement or slack regimes. These efforts will proceed alongside experimental validation on hardware platforms, while end-effector modeling for improved payload handling and extensions to multi-tether configurations will be explored as longer-term research directions.
    \bibliographystyle{IEEEtran}

\bibliography{bibAlias, bibMain, bibAF, bibAR}

\end{document}